\documentclass[letterpaper, 10 pt, conference]{ieeeconf}  

\IEEEoverridecommandlockouts                              

\usepackage[table]{xcolor}
\usepackage{textcomp}
\usepackage{amssymb,amsmath,amsfonts}
\usepackage{graphicx}
\usepackage{tikz}
\usepackage{url}

\usepackage{colortbl}
\usepackage{booktabs}
\usepackage{array}
\usepackage{tabularx}

\newcolumntype{L}[1]{>{\raggedright\let\newline\\\arraybackslash\hspace{0pt}}m{#1}}

\newcommand{\mc}[1]{\ensuremath{\mathcal{#1}}}   

\makeatletter
\DeclareRobustCommand\onedot{\futurelet\@let@token\@onedot}
\def\@onedot{\ifx\@let@token.\else.\null\fi\xspace}

\usepackage[T1]{fontenc}  
\usepackage[b]{esvect}    

\makeatletter
\newlength\xvec@height%
\newlength\xvec@depth%
\newlength\xvec@width%
\newcommand{\xvec}[2][]{%
  \ifmmode%
    \settoheight{\xvec@height}{$#2$}%
    \settodepth{\xvec@depth}{$#2$}%
    \settowidth{\xvec@width}{$#2$}%
  \else%
    \settoheight{\xvec@height}{#2}%
    \settodepth{\xvec@depth}{#2}%
    \settowidth{\xvec@width}{#2}%
  \fi%
  \def\xvec@arg{#1}%
  \def\xvec@dd{:}%
  \def\xvec@d{.}%
  \raisebox{.2ex}{\raisebox{\xvec@height}{\rlap{%
    \kern.05em
    \begin{tikzpicture}[scale=1]
    \pgfsetroundcap
    \draw (.05em,0)--(\xvec@width-.05em,0);
    \draw (\xvec@width-.05em,0)--(\xvec@width-.15em, .075em);
    \draw (\xvec@width-.05em,0)--(\xvec@width-.15em,-.075em);
    \ifx\xvec@arg\xvec@d%
      \fill(\xvec@width*.45,.5ex) circle (.5pt);%
    \else\ifx\xvec@arg\xvec@dd%
      \fill(\xvec@width*.30,.5ex) circle (.5pt);%
      \fill(\xvec@width*.65,.5ex) circle (.5pt);%
    \fi\fi%
    \end{tikzpicture}%
  }}}%
  #2%
}
\makeatother

\makeatletter
\renewcommand*\env@matrix[1][\arraystretch]{%
  \edef\arraystretch{#1}%
  \hskip -\arraycolsep
  \let\@ifnextchar\new@ifnextchar
  \array{*\c@MaxMatrixCols c}}
\makeatother

\renewcommand{\vec}[1]{\xvec[]{#1}}                     
\newcommand{\frm}[1]{\mc{#1}}                           
\newcommand{\point}[1]{{#1}}

\definecolor{commentcolor}{gray}{0.5}
\usepackage{algorithm}
\usepackage{algpseudocode}

\algnewcommand{\LineComment}[1]{\State \textcolor{commentcolor}{\(\triangleright\) #1}}
\algnewcommand{\To}{\textbf{to}}
\algnewcommand{\Break}{\textbf{break}}
\algnewcommand{\Continue}{\textbf{continue}}
\algnewcommand{\IIf}[1]{\State\algorithmicif\ #1\ \algorithmicthen}
\algnewcommand{\EndIIf}{\unskip}
\algnewcommand{\var}[1]{\textit{#1}}
\algnewcommand{\func}[1]{\textsc{#1}}

\newcommand{\OB}{\point{O}_{\frm{B}}}
\newcommand{\OC}{\point{O}_{\frm{C}}}
\newcommand{\OEp}{\point{O}_{\frm{E}}}

\newcommand{\XB}{\hat{x}_b}
\newcommand{\YB}{\hat{y}_b}
\newcommand{\ZB}{\hat{z}_b}
\newcommand{\XC}{\hat{x}_c}
\newcommand{\YC}{\hat{y}_c}
\newcommand{\ZC}{\hat{z}_c}
\newcommand{\XE}{\hat{x}_e}
\newcommand{\YE}{\hat{y}_e}
\newcommand{\ZE}{\hat{z}_e}

\newcommand{\Bframe}{\mathcal{B}}
\newcommand{\Cframe}{\mathcal{C}}

\def \Vbf {\mathbf{V}}
\def \Omgbf {\mathbf{\Omega}}

\def \Rbf {\mathbf{R}}

\def \Tbf {\mathbf{T}}

\usepackage{cite}
\usepackage{soul}
\usepackage{environ}
\usepackage{multirow}
\usepackage[table]{xcolor}
\usepackage{url}
\usepackage{color}

\usepackage{xcolor}
\usepackage{mathtools}
\usepackage{siunitx}
\usepackage{adjustbox}
\usepackage{float}
\usepackage{subcaption}
\usepackage[font=small,tableposition=top]{caption}
\usepackage{booktabs}
\usepackage{bm}

\usepackage{pgfplots}
\pgfplotsset{compat=newest} 
\usetikzlibrary{plotmarks}
\usepgfplotslibrary{patchplots}
\usepackage{grffile}
\usepackage{threeparttable}
\usepackage{gensymb}
\usepackage{todonotes}

\title{\LARGE \bf
Image-based Visual Servo Control for Aerial Manipulation Using a Fully-Actuated UAV
}

\author{Guanqi He$^{1,4*}$, Yash Jangir$^{2,4*}$, Junyi Geng$^{3,4}$, Mohammadreza Mousaei$^{4}$, Dongwei Bai$^{4}$ and Sebastian Scherer$^{4}$
\thanks{$^{*}$ Equal contribution.}%
\thanks{$^{1}$ School of Information Science and Technology, ShanghaiTech University, Shanghai, China. {\tt\small hegq@shanghaitech.edu.cn}}%
\thanks{$^{2}$ Birla Institute of Technology and Science Pilani, Goa Campus, India. 
{\tt\small f20190526@goa.bits-pilani.ac.in}}%
\thanks{$^{3}$ Department of Aerospace Engineering, Pennsylvania State University, University Park, PA, 16802, USA.
{\tt\small jgeng@psu.edu}}%
\thanks{$^{4}$ The Robotics Institute, Carnegie Mellon University, Pittsburgh, PA 15213, USA.
        {\tt\small \{mmousaei, dongweib, basti\}@andrew.cmu.edu}}%
}

\begin{document}

\maketitle
\thispagestyle{empty}
\pagestyle{empty}

\begin{abstract}

Using Unmanned Aerial Vehicles (UAVs) to perform high-altitude manipulation tasks beyond just passive visual application can reduce the time, cost, and risk of human workers. Prior research on \textit{aerial manipulation} has relied on either ground truth state estimate or GPS/total station with some Simultaneous Localization and Mapping (SLAM) algorithms, which may not be practical for many applications close to infrastructure with degraded GPS signal or featureless environments. Visual servo can avoid the need to estimate robot pose. Existing works on visual servo for aerial manipulation either address solely end-effector position control or rely on precise velocity measurement and pre-defined visual visual marker with known pattern. Furthermore, most of previous work used under-actuated UAVs, resulting in complicated mechanical and hence control design for the end-effector. 
This paper develops an image-based visual servo control strategy for bridge maintenance using a fully-actuated UAV. The main components are (1) a visual line detection and tracking system, (2) a hybrid impedance force and motion control system. Our approach does not rely on either robot pose/velocity estimation from an external localization system or pre-defined visual markers. The complexity of the mechanical system and controller architecture is also minimized due to the fully-actuated nature. Experiments show that the system can effectively execute motion tracking and force holding using only the visual guidance for the bridge painting. To the best of our knowledge, this is one of the first studies on aerial manipulation using visual servo that is capable of achieving both motion and force control without the need of external pose/velocity information or pre-defined visual guidance.

\end{abstract}


\section{Introduction} \label{sec:intro}

During the last decade, interest in Unmanned Aerial Vehicles (UAVs) has grown rapidly in a variety of applications, ranging from 3D mapping and photography~\cite{zhao2021super, bonatti2020autonomous}, search and rescue~\cite{moon2022_tigris}, to package delivery with physical interaction~\cite{geng2020cooperative, mousaei2022design, keipour2022physical}. Although UAVs have attracted the interest of researchers, industry, and the general public, most UAV studies continue to focus on passive tasks such as visual inspection, surveillance, monitor and response, remote sensing, etc~\cite{ruggiero2018aerial}.

On the other hand, numerous high-altitude tasks (such as bridge maintenance, wind turbine repairs, and light bulb replacement for high towers) require physical interaction with the environment and are still performed manually. Such hazardous tasks could be automated using UAVs to minimize the risk of human labor as well as reduce time and costs. \textit{Aerial manipulation}) intends to perform manipulation tasks, such as gripping, carrying, assembling, and disassembling mechanical parts, etc. One of the bridges in Pittsburgh, the city of bridges, collapsed at the beginning of year 2022. This is not an isolated incidence; it is part of a pattern in difficult-to-maintain infrastructures like bridges. Employing UAVs to undertake routine maintenance autonomously could help extend the lifespan of such infrastructure.

\begin{figure}
\begin{center}
\setlength{\abovecaptionskip}{0pt}
\includegraphics[width=1.0\linewidth]{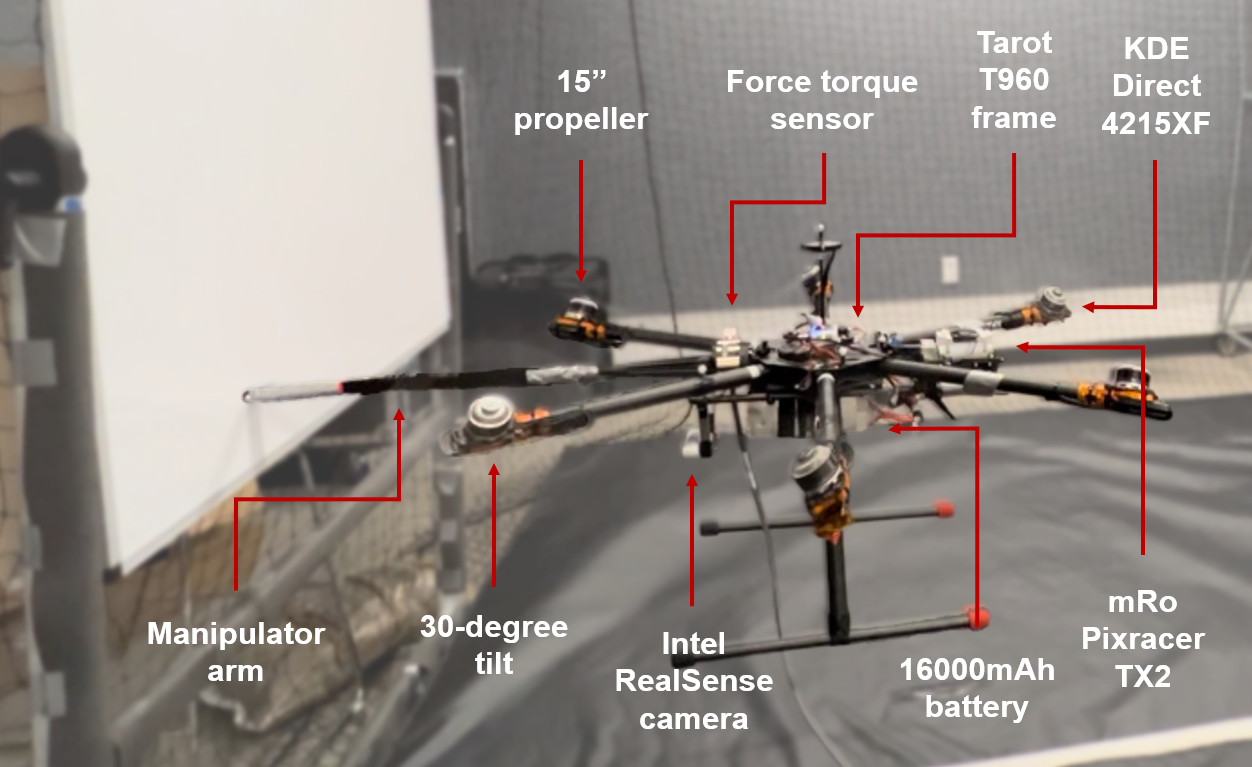}
\end{center}
\vspace{-5pt}
   \caption{A full-actuated UAV performs the painting task.}
   \vspace{-0.7cm}
\label{fig:title}
\end{figure}

There have been many research efforts exploring the aerial manipulation for various kind of jobs~\cite{suarez2020benchmarks}, such as aerial writing~\cite{tzoumanikas2020aerial, lanegger2022aerial}, aerial docking~\cite{choi2022automated}, opening a door~\cite{lee2020aerial}, and pushing a moving cart~\cite{brunner2022energy}, etc. However, the majority of them were studied in a controlled laboratory environment using ground truth state estimation. Although few works explored the outdoor scenario~\cite{hamaza2020design, jimenez2019contact, sanchez2020fully}, they either rely on the GPS or the total station with some SLAM algorithms for localization, which may not be practical in the proximity of the infrastructure due to the degraded GPS signal and featureless environments. This becomes especially critical for aerial manipulation tasks that require high control accuracy because of the coupling nature of the aerial vehicle motion and the manipulation performance. On the other hand, \textit{visual servoing}, especially Image-Based Visual Servoing (IBVS)~\cite{hutchinson1996tutorial}, directly uses the feedback from image coordinates for control~\cite{thomas2014toward}, which can bypass the need for estimating the UAV pose and could be a promising option for aerial manipulation tasks. 

Visual servo control has been widely investigated and commonly used in robotic systems~\cite{espiau1992TRA, andreff2002visual,yu2019siamese, ma2020robotic}. In general, two branches of visual servo approaches are mainly used: position-based (PBVS) and image-based (IBVS), where IBVS does not require precise camera calibration and robot pose estimation compared with PBVS~\cite{hutchinson1996tutorial}. As for UAV application, although the visual servo has been extensively used to assist ship board landing~\cite{keipour2022visual}, target tracking~\cite{zheng2017planning}, etc; applying visual servo for aerial manipulation purposes has been less explored. While few works perform the aerial grasping based on visual servo~\cite{luo2020natural, lai2022image, santamaria2019visual}, most of the application only perform position control on the end-effector, which is not enough for the aerial manipulation tasks that also require precise wrench control, such as holding constant force and torque. 

Researchers recently studied the impedance force control using IBVS for the whiteboard cleaning tasks with pre-defined visual markers~\cite{xu2022image}. However, precise velocity measurement is required from the fusion of motion capture position information and IMU data, which still limits practical usage in the real environment. Because obtaining such precise velocity measurements is challenging, especially in environments with degraded GPS and no visual features. In addition, almost all of the existing work exploring the visual servo for aerial manipulation application use an under-actuated UAV with an additional robotic arm attached to perform manipulation tasks~\cite{zhong2019practical, santamaria2017uncalibrated}, which induces significant complexity on both mechanical and controller design~\cite{ollero2021past}.

This paper develops an IBVS control strategy for bridge maintenance using a fully-actuated UAV. In the scenario of bridge painting, visual servoing becomes especially challenging due to the featureless surfaces. However, these kinds of infrastructure usually contain noticeable edges. We propose a painting strategy that leverages only the original edges of the bridge and the self-painted lines in the process for visual guidance. 
The system consists of two major components: a visual line detection and tracking system and a hybrid motion and impedance force control system. The former detects the edges of the bridges as well as the painted lines, then continues tracking them to provide visual guidance for the control system. The latter enables the aerial manipulator to maintain constant pushing force while conducting the lateral motion. 
Our approach does not rely on either robot pose/velocity estimation from an external localization system or pre-defined visual guidance such as markers with known pattern.


In summary, the main contributions of this work are:
\begin{itemize}
    \item We develop an image-based visual servo control strategy for bridge maintenance application using a fully-actuated UAV. Our approach does not rely on either robot pose/velocity estimation from an external localization system, such as GPS/motion capture system, or the pre-defined visual markers.
    
    \item We develop a hybrid motion and impedance force controller so that the aerial manipulator can maintain constant force while tracking the lateral motion. Benefiting from the fully-actuated UAV platform, the complexity of the controller design gets reduced significantly. 
    
    \item We design an efficient line detection and tracking algorithm, which leverages the surface normal to provide the filtered surface. Our method can be run at a high computation rate, which is critical for real-time usage. 

    \item We present simulation and experiments to evaluate the whole motion tracking and force holding performance under visual guidance. 
\end{itemize}

\section{Overview} \label{sec:strategy}

\begin{figure}
\begin{center}
\setlength{\abovecaptionskip}{0pt}
\includegraphics[width=1.0\linewidth]{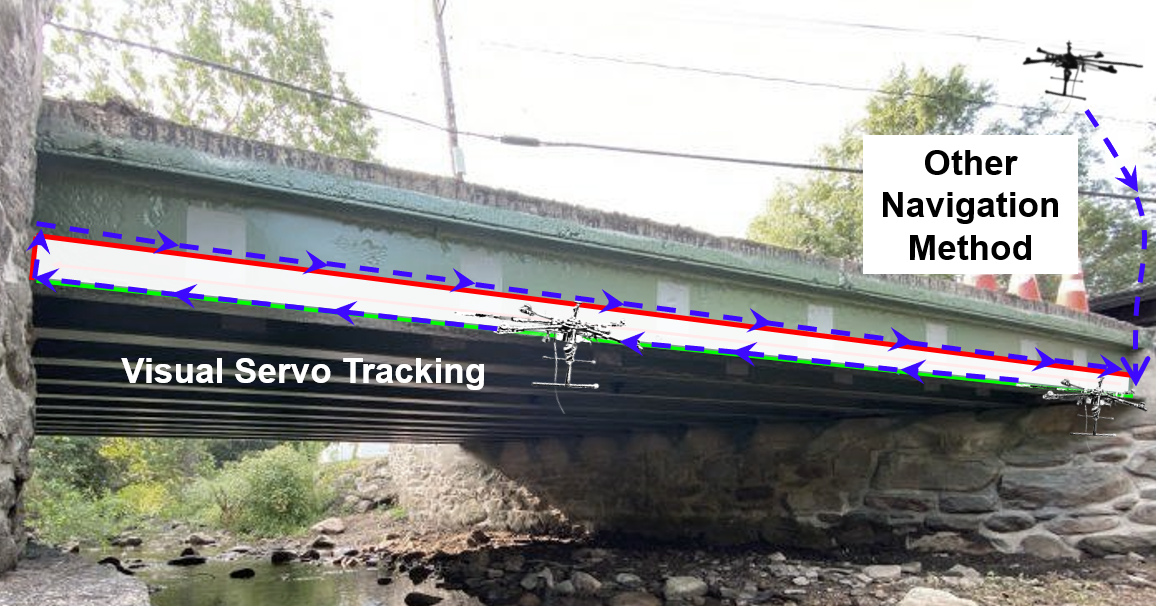}
\end{center}
\vspace{-5pt}
   \caption{Illustration of bridge painting strategy.}
   \vspace{-0.5cm}
\label{fig:painting}
\end{figure}

\subsection{System Overview}\label{sec::system}

Here we discuss our vehicle platform and briefly describe its mechanical design and avionics.

\subsubsection{Vehicle Design}

As shown in Figure~\ref{fig:title}, our vehicle is a hexarotor (Tarot T960) with all rotors tilted 30 degrees with interleaving left and right tilt directions. Each arm (totaling six) has a brushless electrical motor (KDE Direct 4215 XF) that can drive a $15"$ propeller and deliver $52.56$N at full throttle. Thanks to the fully-actuated nature, a zero degree-of-freedom (DoF) manipulator arm is attached to the front of the vehicle without extra complex mechanical component. A 6 DoF force-torque sensor is attached to the based of the manipulator to measure the forces and moments.

\subsubsection{Avionics}

The flight controller on our vehicle is a mRo Pixracer (FMUv4). It features 180 MHz ARM Cortex® M4 processor, inertial measurement unit (IMU), barometer, and gyroscope. This flight controller hardware uses our version of customized PX4 firmware \cite{keipour20}, allowing it to control our fully actuated platform. Additionally, onboard computation is running on an Nvidia Jetson TX2 equipped with dual-core Nvidia CPU and Nvidia Pascal GPU. Finally, we employed an Intel RealSense depth camera mounted under the manipulator arm as the vision sensor.

\begin{figure*}
\centering
\includegraphics[width=0.9\linewidth]{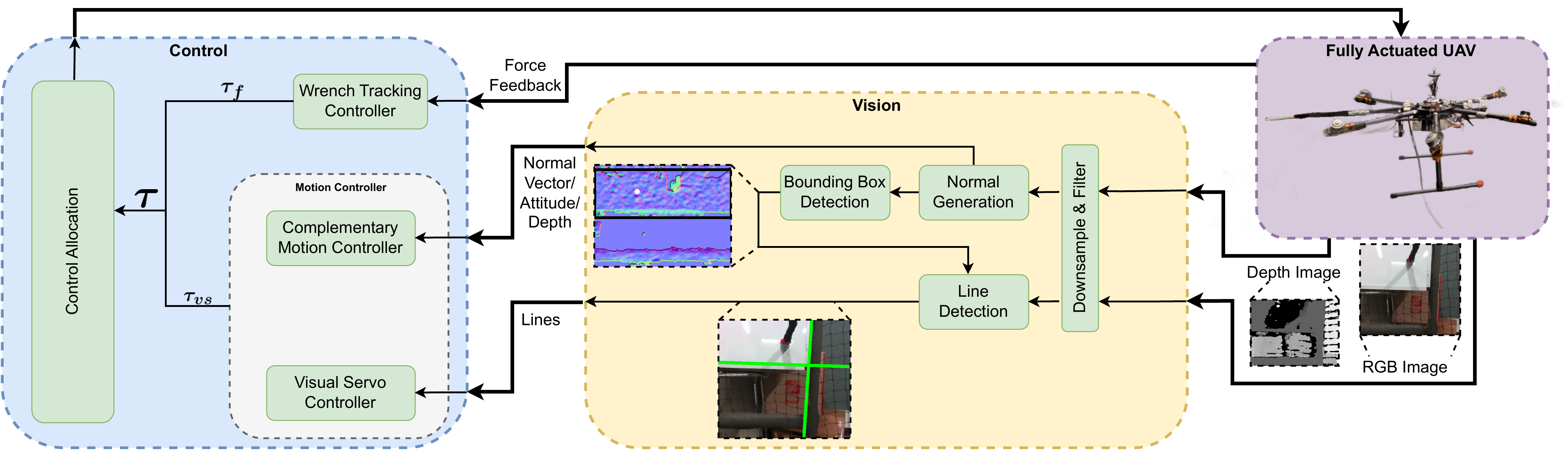}
\caption{Architecture of the vision-guided hybrid motion and impedance controller}\label{diagram}
\vspace{-0.5cm}
\end{figure*}

\subsection{Strategy Overview}

This section presents an overview of our bridge painting strategy. We first make two assumptions: (1) bridges usually have edges that can be leveraged for visual guidance, such as in the scenario shown in Figure~\ref{fig:painting}. The edges are typically horizontal and vertical; (2) when the aerial vehicle is far away from the infrastructure, other guidance methods can be used, such as GPS. Based on these assumptions, the vehicle first flies close to the bridge to a pre-selected starting point, such as the bottom right corner, under other guidance methods. Then, we switch to visual servo control by detecting and tracking the bottom edge (green line in Figure~\ref{fig:painting}) and perform the painting from right to left. As the robot reaches the other side of the bridge, it starts tracking the vertical edge to move up and then switches to the lateral direction to paint back. During the painting process, the newly painted line generates new edges (horizontal red line), which serve as visual guidance for continuous painting.

Notice that we only leverage the visual line feature during the painting process without relying on external guidance to estimate vehicle pose or velocity. This indicates that the system needs a reliable line detection and tracking module to provide real-time visual feature. In addition, to satisfy the mission requirement -- good painting quality in this scenario, the aerial manipulator needs to hold a constant force in the orthogonal direction of the bridge surface while maintaining an effective lateral motion (or vertical motion in the moving up phase).

\section{Image-based Visual Servo Control} \label{sec:visual-servo} 

This section describes the image-based visual servo control for tracking lines in the bridge painting tasks. We first define the notations and introduce the multi-rotor dynamics for the fully-actuated UAV. Then, we develop the hybrid motion and force control system,
so that the aerial manipulator can hold constant pushing force while maintaining an effective motion to guarantee the painting quality.

\subsection{Notations}
Four reference frames are defined: world inertial frame~$\mathcal{I}$, vehicle body frame~$\mathcal{B}$, camera frame~$\mathcal{C}$, and end-effector frame~$\mathcal{E}$. The inertial frame is defined as the north-east-down frame with the origin to be the initial contact point of the UAV to the surface. The body frame is defined as $\mathcal{B} = \{\OB, \XB, \YB, \ZB\}$, where $\OB$ is the position of the vehicle's center of mass, and $\XB$, $\YB$, and $\ZB$ are the unit vectors pointing to the front, right and bottom directions of the vehicle, respectively. The camera frame is defined as $\mathcal{C} = \{\OC, \XC, \YC, \ZC\}$, where $\OC$ is the position of the camera optical center, and $\XC$, $\YC$, and $\ZC$ are the unit vectors pointing to the right, down and front directions of the camera, respectively. The end-effector frame is defined as $\mathcal{E} = \{\OEp, \XE, \YE, \ZE\}$, where $\OEp$ is the tip of the end-effector that will contact with the other surface, and $\XE$, $\YE$, and $\ZE$ are the unit vectors pointing to the right, up and backward directions of the contact point, respectively. 

$\Rbf_a^b \in \mathbb{R}^{3\times3}$, $\Tbf_a^b\in \mathbb{R}^{6\times6}$ define the and rotation matrix and the twist transformation matrix from frame $a$ to $b$, respectively. For the convenience, we also denote $e_i$ as a unit vector with the $i^\mathrm{th}$ component as 1, $\mathbf{0}_{n\times n}$ as the $n\times n$ zero matrix, $\mathbf{I}_{n\times n}$ as the $n\times n$ identity matrix.


\subsection{Multi-rotor Dynamics}

The dynamic model of the hexarotor aerial manipulator can be derived using the Lagrangian method~\cite{bodie2021TRO}:
\begin{align}
\label{drone-model}
    \bm{M \dot v + C v  = \tau' + G}
\end{align}
with inertia matrix $\bm{M}\in\mathbb{R}^{6\times 6}$, centrifugal and Coriolis term $\bm{C}\in\mathbb{R}^{6\times 6}$, gravity wrench $\bm{G}\in\mathbb{R}^{6}$, control wrench $\bm{\tau'}\in\mathbb{R}^{6}$ and $\bm{v} = \begin{bmatrix}
    \Vbf^\top\  & \Omgbf^\top
\end{bmatrix}^\top \in \mathbb{R}^{6}$ system twist (linear and angular velocity) expressed in the body frame $\Bframe$, 
\begin{align}
    \bm{M} & = \text{diag}\left(\begin{bmatrix}
        m\mathbf{I}_{3\times3} & \bm{J}
    \end{bmatrix}\right)\notag\\
    \bm{C} & = \text{diag}\left(\begin{bmatrix}
        m[\Omgbf]_\times & -\bm{J}[\Omgbf]_\times
    \end{bmatrix}\right)\\
    \bm{G} & = -m\text{diag}\left(\begin{bmatrix}
        \mathbf{R}_{\mathcal{I}}^{\mathcal{B}} & \mathbf{0}_{3\times3}
    \end{bmatrix}\right)\bm{g}\notag  
\end{align}
where $m$ is the vehicle mass, $\mathbf{J}$ is the moment of inertia, $\mathbf{g}$ is the gravity. $[\bm{*}]_\times$ is the skew-symmetric matrix associated with vector $\bm{*}$. 

Since the system is fully-actuated, we apply the feedback linearization input $\bm{\tau}'=\bm{C} \begin{bmatrix}
    \mathbf{0}_3 \\ \Omgbf
\end{bmatrix} - \bm{G} + \bm{\tau}$. Note that only the angular velocity and attitude is required in this compensation, which can be obtained from the IMU sensor. Because it is hard to obtain reliable velocity measurement during the bridge painting when the UAV is close to the infrastructure, we set the velocity compensation to be $\mathbf{0}_3$. The dynamics of the system \eqref{drone-model} then becomes \begin{align}\label{dynamics-linearized}
    &\bm{M}\bm{\dot v} = \bm{\tau} + \bm{\tau}_{cor}
\end{align} 
with $\bm{\tau}_{cor}=-\begin{bmatrix}
        m \bm{[\Omgbf]_\times \Vbf} \\ \mathbf{0}_3
    \end{bmatrix}$.


\subsection{Hybrid Motion and Force Control}
We leverage the advantages of the fully-actuated UAV to design the hybrid motion and force controller. Different from the traditional coplanar multi-rotors, which can only generate thrust normal to its rotor plane, requiring it to completely tilt towards the total desired thrust direction to align the generated thrust with the desired thrust, the fully-actuated vehicles are capable of independently controlling their translation and orientation. Through the control input $\bm{\tau}$, either the motion tracking tracking controller $\bm{\tau}_{vs}$ (here mainly driven by visual servo) or the wrench tracking controller $\bm{\tau}_f$ can be implemented as~\cite{bodie2021TRO} \begin{align}
    \bm{\tau} &= (\mathbf{I}_{6\times 6}-\bm{\Lambda}) \bm{\tau}_{vs} + \bm{\Lambda} \bm{\tau}_f \label{eq:hybrid}\\
    \bm{\Lambda} &= \text{blockdiag}(\bm{\Lambda}', \mathbf{0}_{3\times 3})\in\mathbb{R}^{6\times 6}\\
    \bm{\Lambda}' &= \mathbf{R}_{\mathcal{E}}^\Bframe\begin{bmatrix}
        0 & 0 & 0\\
        0 & 0 & 0\\
        0 & 0 & \lambda(d)\\
    \end{bmatrix}\label{eq:selection}
\end{align} with {$\lambda(d)\in[0,1]$, $d$ the depth measurement of the camera between the UAV and the contact surface. The matrix $\mathbf{\Lambda}$ selects direct force control commands, and leaves the complementary subspace for the motion control. Since the end-effector is a single rigid link, only the elements in the $\mathbf{\Lambda}$ that correspond to the direction of pushing force are set to $1$, and the other elements are set to 0. In the end, the control wrench $\bm{\tau}$ is mapped to the rotor speeds of the fully-actuated UAV through the control allocation process\cite{keipour2022physical}.

\subsubsection{Impedance Wrench Tracking Controller}
In reality, we noticed in the experiment that direct force control leads to significant oscillation and cannot absorb the pushing energy efficiently. Therefore, we designed an impedance force control scheme to ensure the end-effector holds a constant reference force ${F}_{ref}$ during the painting. The interaction force ${F}_{t}$ acting on the UAV body frame is measured by an onboard force and torque(F/T) sensor. The force-tracking control action is computed as 
\begin{align}
    {e}_f\! &= {F}_t - {F}_{ref}\notag\\
    {F}_{f}\! &=\! {F}_{ref} - m m_d^{-1} ({K}_s {e}_s + {D}_s \dot{e}_s)+{K}_{f,p}{e}_f + {K}_{f,i}\int{e}_f dt\notag\\
    &\!= {F}_{ref} + {K}_{s,p} {e}_s + {K}_{s,d} \dot{e}_s+ {K}_{f,p}{e}_f + {K}_{f,i}\int{e}_f dt
\end{align} 
with ${K}_{f,p}$, ${K}_{f,i}>0$ the tunable gains, $m, m_d$ the actual and desired vehicle mass, respectively. $K_{s,p}=m m_d^{-1} K_s$ is the normalized stiffness. $K_{s,d}=m m_d^{-1} D_s$ is the normalized damping. Then, the control wrench $\bm{\tau}_f = \begin{bmatrix}
       {F}_{f} & 0 & 0 & 0 & 0 & 0
\end{bmatrix}^\top\in\mathbb{R}^6$ will be pass through (\ref{eq:hybrid}) for the hybrid motion and wrench control.

Actually in (\ref{eq:selection}), $\lambda(d)$ represents the selected weight of hybrid modes, where $\lambda(d)=0$ corresponds to the pure motion control, $\lambda(d)=1$ indicates the impedance force control along the normal direction of the wall. The wrench controller is activated when the aerial manipulator is getting contact with the working surface. We designed $\lambda$ as a confidence factor that gradually increases with depth $d$.
\begin{align}
    \lambda(d)=\left\{ \begin{array}{lll}
      1, & \text{if $d\leq d_{min}$} \\
      \frac{1}{2}(1+\cos{\frac{d-d_{min}}{d_{max}-d_{min}}}\pi), & \text{if $d_{min}<d\leq d_{max}$}\\
      0, & \text{otherwise.}
    \end{array} \right.
\end{align}
The whole wrench control operates as the impedance controller to dynamically control the related force and motion.

\subsubsection{Visual-servoing Motion Controller}
In this section, we design a line-based IBVS controller to ensure the motion of the end-effector is aligned with the tracking line obtained from the detected bridge or painted edge.


Let $\bm{q}=[{\bm{q}'_1}^\top,\cdots, {\bm{q}'_n}^\top]^\top\in\mathbb{R}^{2n}$ be the image feature vector, where $\bm{q}'_i=[\rho_i, \theta_i]^\top$ denotes the $i^\mathrm{th}$ image feature: the Hough parameter\footnote{The straight can be represented by the slope-intercept. However, vertical lines pose a problem. They would give rise to an unbounded slope. The Hough space representation avoids this issue.} of the $i^\mathrm{th}$ line on the image. $n$ is the total number of lines detected on the image. The image feature error represents the error between the desired and actual position of image features $\bm{e}_q = \bm{q} - \bm{q}_{ref}$. Then, the first-order error dynamics can be represented as: 
\begin{align}\label{IBVS-velocity}
    \dot {\bm{e}}_q &= \dot {\bm{q}} - \dot {\bm{q}}_{ref} =\bm{L}(q)\bm{v} -  \dot {\bm{q}}_{ref}
\end{align} 
with $\bm{L}(q) = \bm{L}_c(q) \bm{T}^\Cframe_\Bframe$, $\bm{L}_c\in \mathbb{R}^{2n\times 6}$ the interaction matrix for multiple-line visual features. A 3D line can be defined by the intersection of two planes~\cite{chaumette1993classification}
\begin{align}
    \left\{ \begin{array}{cc}
         A_1X+B_1Y+C_1Z+D_1&=0\\
         A_2X+B_2Y+C_2Z+D_2&=0
    \end{array} \right.
\end{align} with $D_1^2+D_2^2\neq 0$.
Then interaction matrix of a line on image with $(\rho,\theta)$ is written as 
\begin{align}\label{eq-line-jacobian}\small
    \bm{L}({q})=\begin{bmatrix}
    \lambda_\theta c_\theta & \lambda_\theta s_\theta & -\lambda_\theta \rho & -\rho c_\theta & -\rho s_\theta & -1\\
    \lambda_\rho c_\theta & \lambda_\rho s_\theta & -\lambda_\rho \rho & (1+\rho^2) s_\theta & -(1+\rho^2) c_\theta & 0\\
    \end{bmatrix}
\end{align}
The terms $c_\theta$, $s_\theta$ refer to $\cos(\theta)$ and $\sin(\theta)$ respectively. The scalars $\lambda_\theta$ and $\lambda_\rho$ are given by \begin{align}
    \lambda_\theta&=(A_i \sin\theta-B_i \cos\theta)/D_i\notag\\ \lambda_p&=(A_i\rho\cos\theta+B_i\rho\sin\theta+C_i)/D_i \notag
\end{align} with $A_i$, $B_i$, $C_i$, and $D_i$ the parameters of the $i^\mathrm{th}$ planes defining the 3D line of interest. Then, the interaction matrix for multiple lines can be represented as the stack of each single-line interaction matrix. \begin{align}
    \bm{L}_c(q) = \begin{bmatrix}
        \bm{L}(q'_1)\\
        \vdots\\
        \bm{L}(q'_n)
    \end{bmatrix}
\end{align}

By differentiating \eqref{IBVS-velocity}, the image space error dynamics can be obtained as \begin{align}
    \ddot {\bm{e}}_q = \dot{\bm{L}}\bm{v} + \bm{L}\bm{M}^{-1} \left( \bm{\tau}_{vs} + \bm{\tau}_{cor} \right)-\ddot {\bm{q}}_{ref}
\end{align}
In the bridge painting scenario, $\bm{q}_{ref}$ is a piecewise static visual target trajectory with $\ddot{\bm{q}}_{ref}=0$. The visual servoing controller can then be designed as:
\begin{align} \label{vs-v1}
    \bm{\tau}_{vs} = - \bm{M}\hat{\bm{ L}} ^\dagger \left( \bm{K}_{q,p} \bm{e}_q + \bm{K}_{q,d} \dot{\bm{e}}_q +\bm{K}_{q,i} \int{\bm{e}}_q dt \right)
\end{align}
where $\hat{\bm{L}}$ is the approximation of the interaction matrix $\bm{L}$ by using the line approximation at the desired position, $\hat{\bm{ L}} ^\dagger$ is the pseudo-inverse of $\hat{\bm{L}}$, and $\bm{K}_{q,p}$, $\bm{K}_{q,d}$, $\bm{K}_{q,i}\in\mathbb{R}^{6\times 6}$ are diagonal and positive definite matrices. It can be easily proved that for a bounded system twist, $\bm{\tau}_{cor}$ can be controlled by properly designing the controller gains in (\ref{vs-v1}). In fact, $\bm{\tau}_{cor}\leq \gamma \|\Omgbf\| v$.


Note that the dimension of $\bm{L}$, $\hat{\bm{L}}$ and $\hat{\bm{ L}} ^\dagger$ varies with the number of detected lines in the image. In general, three non-parallel lines can completely constrain the vehicle motion. In the scenario of bridge painting, the edges of bridge or the painted lines are the potential candidates to be detected. Due to the limited camera field of view, for most of the time during the painting, either two lines (one horizontal line: bottom or painted edge, and one vertical edge line) or only one horizontal line (bottom or painted edge) can be observed. 

When both one horizontal and one vertical line are available or $n=2$, we have $\bm{q} = \begin{bmatrix}\rho_1,\theta_1,\rho_2,\theta_2\end{bmatrix}^\top \in\mathbb{R}^4$, and the desired configuration $\hat{\bm{q}} =\begin{bmatrix}
    f_y{l_1}/{d_1}& {\pi}/{2}& f_x{l_2}/{d_2}& 0
\end{bmatrix}$ with $f_x$, $f_y$ to be the camera focus length, $d_i$ to be the relative depth of the $i^{\mathrm{th}}$ line and $l_i$ is offset of the $i^{\mathrm{th}}$ line with respect to the camera optical axes in the image plane. Define each of the two lines as the intersection of a horizontal and a vertical plane, the equations w.r.t the camera frame are given by \begin{align}
    \left\{\begin{array}{cc}
         Y-l_1&=0  \\
         Z-d_1&=0 
    \end{array} \right. \quad     \left\{\begin{array}{cc}
         X-l_2&=0  \\
         Z-d_2&=0 
    \end{array} \right.
\end{align} This gives the interaction matrix 
\begin{align}
    \bm{\hat{L}}_c(\bm{q}) = \begin{bmatrix}
        0 & 0 & 0 & 0 & -\frac{f_y l_1}{d_1} & -1\\
        0 & -\frac{1}{d_1} & -\frac{f_y l_1}{d_1^2} & 1+\frac{f^2_yl_1^2}{d_1^2} & 0 & 0\\
        0 & 0 & 0 & -\frac{f_x l_2}{d_2} & 0 & -1\\
        -\frac{1}{d_2} & 0 & -\frac{f_x l_2}{d_2^2} & 0 &-1-\frac{f^2_xl_2^2}{d_1^2} & 0
    \end{bmatrix}
\end{align}

If only one horizontal line is detected on the image or $n=1$, $\bm{q} = \begin{bmatrix}\rho_1,\theta_1\end{bmatrix}^\top \in\mathbb{R}^2$ and $\hat{\bm{q}} =\begin{bmatrix}
    f_y{l_1}/{d_1}& {\pi}/{2}
\end{bmatrix}$. Similarly, the interaction matrix can be written as \begin{align}
    \bm{\hat{L}}_c(\bm{q}) = \begin{bmatrix}
        0 & 0 & 0 & 0 & -\frac{f l_1}{d_1} & -1\\
        0 & -\frac{1}{d_1} & -\frac{f_y l_1}{d_1^2} & 1+\frac{f^2_yl_1^2}{d_1^2} & 0 & 0
    \end{bmatrix}
\end{align}

\subsubsection{Complementary Motion Controller}
In both cases of $n=1, 2$, $\hat{\bm{ L}} ^\dagger$ has a non-empty null space, which indicates that the UAV motion cannot be fully determined by only using the visual servoing control strategy. Additional control mechanism is needed to further constrain the vehicle motion. Therefore, by referring to the control scheme in \cite{chaumette2010IROS}, we designed an extra complementary motion controller $\bm{\tau}_p$, then the overall motion controller can be expressed as:
\begin{align} \label{vs-v2}
    \bm{\tau}_{vs}\! =\! - \bm{M}\!\left[\!\hat{\bm{ L}} ^\dagger \!\left(\! \bm{K}_{q,p} \bm{e}_q \!+ \!\bm{K}_{q,d} \dot{\bm{e}}_q \!+\! \bm{K}_{q,i} \!\int \! {\bm{e}}_q dt \right) \!+ \!\bm{P}_{vs} \bm{\tau}_p\right]
\end{align}
where $\bm{P}_{vs} = (\mathbf{I}_{6\times6}- \hat{\bm{ L}} ^\dagger \hat{\bm{ L}})$ is a projection operator on the null space of $\hat{\bm{ L}}$ so that the complementary motion controller $\bm{\tau}_p$ can be achieved at the best under constraint without perturbing the regulation of $\bm{e}_q$ to be $\mathbf{0}$. The purpose of $\bm{\tau}_p$ here is to (1) control the attitude of the vehicle to keep level flight thus facilitates the control of the pushing force; (2) assist in controlling the vehicle lateral motion. 

Let $\bm{\tau}_p = \begin{bmatrix}
     \bm{F}_p^\top & \bm{M}_p^\top
 \end{bmatrix}^\top$. Thanks to the fully-actuated nature of the UAV which provides more controllable degree of freedom so that the vehicle translation and orientation can be controlled independently. We leverage our previous work~\cite{keipour2022physical} and select the zero-tilt (zero roll and pitch) attitude strategy to keep the vehicle tilt at zero all time and stay completely horizontal during the flight. Because the painting quality is affected by the pushing force and the lateral motion of the vehicle, keeping the manipulator level can enable precise contact with the vertical surface. 
As for the yaw control, we use the normal vector $\mathbf{n}$ of the contact surface obtained from the RGB-D camera as the attitude feedback, so that the body frame is always aligned with the wall using proper camera feedback. The overall complementary control torque $\bm{M}_p$ can thus be obtained.

In fact, under the zero-tilt strategy, when $n=2$ with two non-parallel lines in the camera field of view, both vertical and lateral motion can be controlled by the visual servo controller. In the bridge painting scenario, the vertical edges of the bridge provides visual guidance to terminate the lateral motion and enable the painting direction switch.
In the phase of lateral painting, when $n=1$ with only a horizontal line, visual servo controller only provides feedback to the vertical motion of the vehicle. It is challenging to control the lateral motion since the detected line does not provide vehicle feedback. We instead estimate the vehicle velocity from the optical flow of the sparse features and the IMU information, then use it for the lateral motion controller design. 

Let us define the estimated velocity $V_x$ and the lateral velocity target $V_d$, the error $e_v = V_x - V_d$, the lateral motion controller under single horizontal line scenario with zero-tilt strategy is designed as: 
\begin{align}
     F_x =  K_{x,p} e_v + K_{x, i} \int e_v dt
\end{align}
where $K_{x,p}$ and $K_{x,i}$ are control gains. $\bm{F}_p = \begin{bmatrix}
       0 & F_x & 0
\end{bmatrix}^\top$.
Thanks to overall hybrid control strategy, the estimated velocity can enable the lateral motion to be controlled effectively. Actually, in reality, besides the sparse features, we can also actively design the painting patterns, such as alternate horizontal and vertical paths to generate extra visual guidance for lateral motion control.

\section{Line Detection and Tracking}\label{sec:line}
In the case of line detection and tracking in a real scenario, the main challenge comes from how to accurately reject lines in background noise and only maintain the major line of the bridge edge. To tackle this challenge, we leverage the fact that the vehicle is always operating on the same surface and perform the filtering based on the surface normal. Our algorithm can be run at a high computation rate, which is critical for real-time usage. 

The Real-sense camera provides both RGB and depth images. The overall algorithm consists of three main components: (1) Normal Image generation; (2) Bounding Box estimation; (3) Segmentation and Line detection. 

\subsection{Normal Image generation}
After pre-processing, we compute the surface normal of the filtered depth image. For a 3D point ${}^{\Cframe}(x, y, z)$ with the pixel coordinate $(u, v)$, the normal vector on this point can be computed by the gradient of the depth image:

\begin{align}
    \vec{n}^{\prime} &= \left(\frac{\partial z}{\partial x}, \frac{\partial z}{\partial y}, 1\right), \quad
    {}^\Cframe\mathbf{n} =\frac{\vec{n}^{\prime}}{\left\|\vec{n}^{\prime}\right\|} 
\end{align}
where,
\begin{align}
    \frac{\partial z}{\partial x} &= \frac{\partial z}{\partial u} \cdot \frac{f_{x}}{z} \approx \frac{\Delta z}{\Delta u} \cdot \frac{f_{x}}{z}, \quad
    \frac{\partial z}{\partial y}=\frac{\partial z}{\partial v} \frac{f_{y}}{z} \approx \frac{\Delta z}{\Delta v} \cdot \frac{f_{y}}{z}
\end{align}

\subsection{Bounding Box estimation}
From an initial point on the surface normal, we keep searching over different directions until high order change happens under some threshold, representing the boundary of the target surface under operation. Then, a bounding box is generated around the center as the initialization point for the new image frame. 

\subsection{Segmentation and Line detection} 
The bounding box is then used to segment the surface from a down-sampled RGB image. Then, the line detection is performed only in the segmented image to reduce computation time. Specifically, we use the canny edge detection algorithm, and the probabilistic Hough transform to detect the major lines on the surface. Then, thresholding based on line slope is performed to extract the vertical and horizontal lines, which are then provided as visual guidance to the controller. In the bridge painting scenario, we always focus on the topmost horizontal line for lateral tracking or the first seen vertical line for direction switching.

\section{Experiments and Results} \label{sec:results}

\subsection{Experiment Setup}
We modeled our fully-actuated aerial manipulator in the Gazebo simulator based on the design described in Sec. \ref{sec::system}. The system runs on Robot Operating System (ROS) in Ubuntu 18.04. The system software is developed upon CMU Airlab's core autonomy stack with the PX4 firmware as the control stack. An RGB-D camera is also modeled with the same camera parameters as the Real-Sense and is attached to the front of the aerial manipulator. In addition, a 6-axis force and torque sensor Gazebo plugin model is added to the upper front of the vehicle. 

As for the environment, we create a wall with a whiteboard surface on top of it for the manipulator to perform the painting tasks. In a real experiment, we notice that the surface property and the friction force significantly affect the contact behavior of the vehicle and the general visual servo tracking performance. Therefore, we add the Coulomb friction model to the surface. 

\subsection{Results}
\subsubsection{Force Tracking Evaluation}
This experiment verifies the force tracking and the corresponding visual-seroving performance. The aerial manipulator first stabilizes at a close contact position with the surface edge in the camera's field of view. Once the visual servo and force control is activated, the vehicle moves to the visual target vertically and maintain the desired force 5N.

The results are summarized in Figure~\ref{fig:impedance}, which shows the force tracking and the visual target tracking performance. The vehicle starts force-holding and stabilizes around 5N once the force control and visual servo is activated. It takes about 10s for the aerial manipulator to exert the external force around the desired value. We notice that at about 20s, the vehicle gets in contact with the surface, which generates sudden disturbance to the visual tracking. During the period of 20s to 30s, the impedance force controller is transitioning to the target force, which generates another disturbance to the visual tracking. In fact, during this period, the vehicle zero-tilt is also in transition, which causes a slight non-zero tilt of the vehicle pose, resulting in a disturbance to the visual 

\begin{figure}
	\centering
	\subfloat[\label{fig:imp}]{\includegraphics[width=0.48\linewidth]{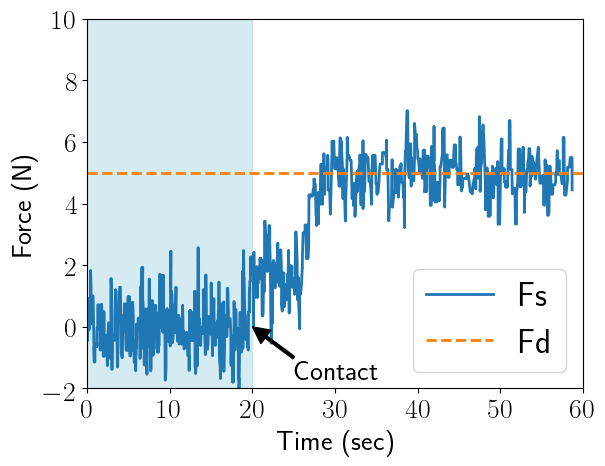}}
	\subfloat[\label{fig:pure}]{\includegraphics[width=0.48\linewidth]{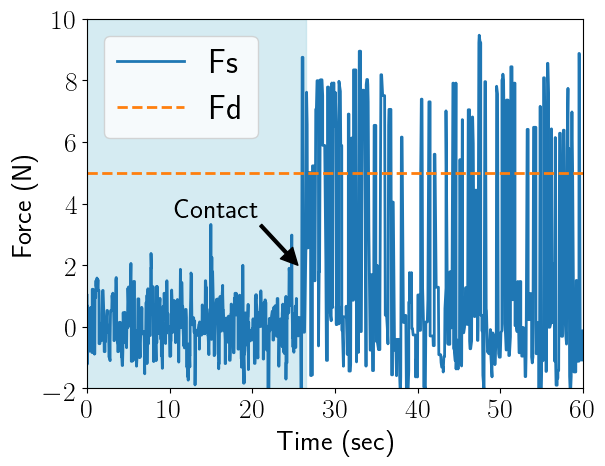}} \\
	\subfloat[\label{fig:imp_feature}]{\includegraphics[width=0.48\linewidth]{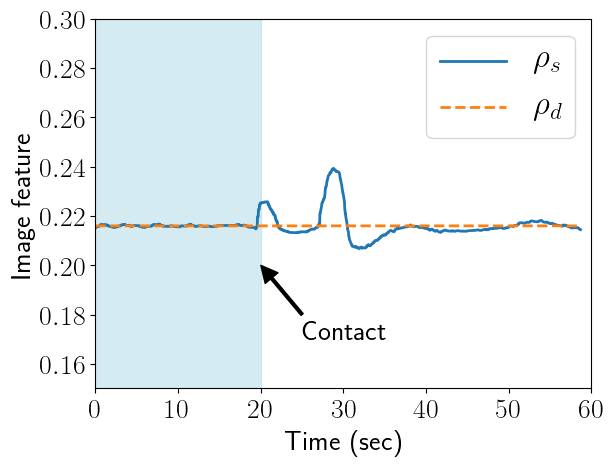}}
	\subfloat[\label{fig:pure_feature}]{\includegraphics[width=0.48\linewidth]{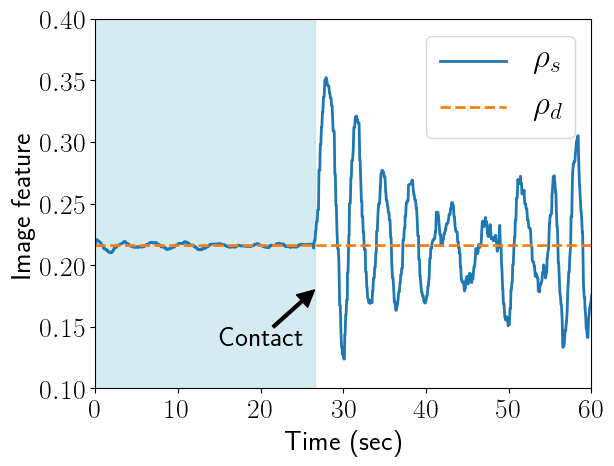}} \\
	\caption{Experimental results of the aerial manipulator tracks a desired force. (a) and (b) external force. (a) results from our impedance force controller. (b) results from pure force control. (c) and (d) visual target tracking for the two methods.}
	\label{fig:impedance}
	\vspace{-15pt}
\end{figure}
\noindent tracking. This disturbance reaches a peak at about 30s and is then compensated by the visual servo controller. 

We compare our method with the direct force control, as shown in Figure~\ref{fig:pure}. We can see that without impedance, the force tracking runs into significant oscillation, which leads to the final crash in the real hardware test. The reason is that adding the impedance component effectively absorbs the pushing energy and avoids the energy divergence.

\begin{figure}

	\centering
	\subfloat[\label{fig:C3D}]{\includegraphics[width=0.48\linewidth,height=0.42\linewidth]{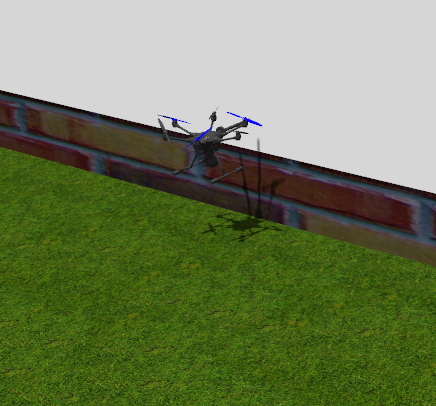}}
	\subfloat[\label{fig:C3D}]{\includegraphics[width=0.48\linewidth]{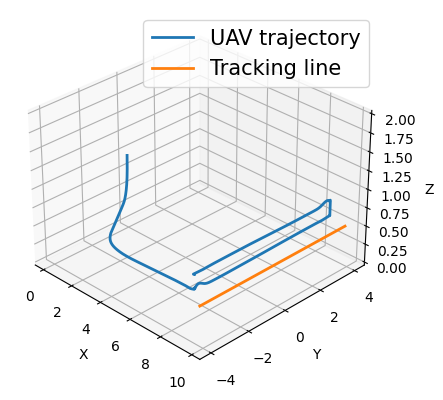}}\\
	\subfloat[\label{fig:Cforce}]{\includegraphics[width=0.48\linewidth]{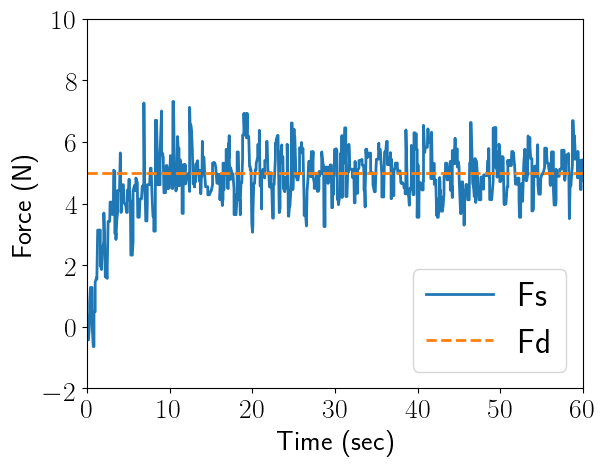}} 
	\subfloat[\label{fig:Cfeature}]{\includegraphics[width=0.48\linewidth]{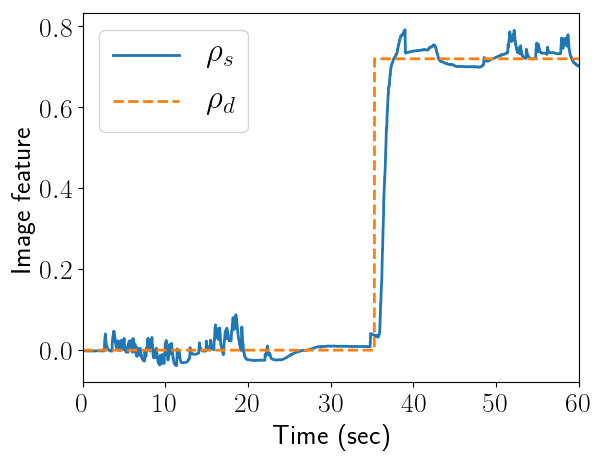}}\\
	\subfloat[\label{fig:Cvel}]{\includegraphics[width=0.48\linewidth]{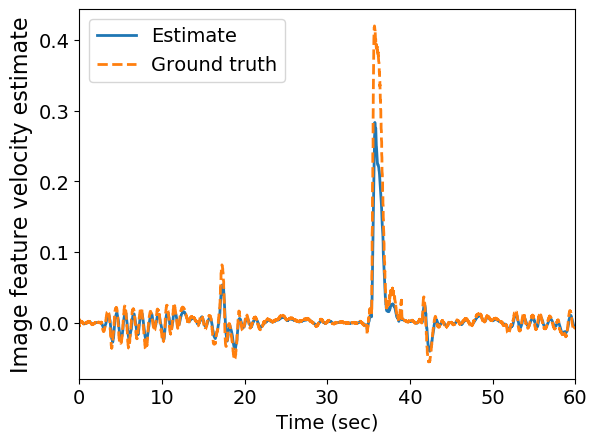}} 
	\subfloat[\label{fig:tilt}]{\includegraphics[width=0.48\linewidth]{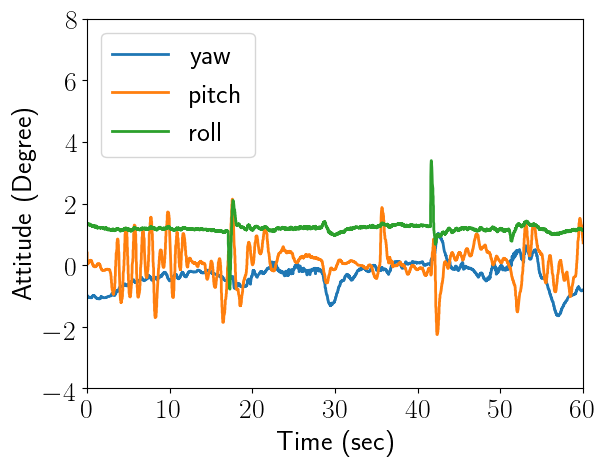}} \\
	\caption{Experimental results of the aerial manipulator conducts a complete demonstrations. (a) a snapshot of the experiment. (b) vehicle 3D trajectory. (c) external force. (d) visual target tracking. (e) image plane velocity estimation. (f) vehicle attitude.}
	\label{fig:Cshape}
	\vspace{-15pt}
\end{figure}

\subsubsection{Lateral Motion with Force Holding}\label{sec:demo}
We then present a complete demonstrations of the overall bridge painting processes using the developed visual servo control system. 
Figure~\ref{fig:Cshape} presents the overall performance. The aerial manipulator is able to track the visual line while maintaining the pushing force although without pose estimate or external velocity information. At about 35s, the vehicle reaches the left boundary of the surface then switches to vertical line tracking followed by a lateral back tracking. Notice that there is a big spike of the estimated velocity followed by a small spike at about 43s, which is another direction changing. From Figure~\ref{fig:tilt}, it is clear that the vehicle is keeping zero tilt all the time. 
We also present line detection and tracking performance in a real environment shown in Figure~\ref{fig:realline}. We can see our algorithm accurately detects surface edges. Algorithm runs on Intel i7 3.6GHz CPU laptop providing line detection at 27Hz and achieves 12Hz with Nvidia Jetson TX2 utilizing only onboard dual-core CPU.


\subsubsection{Noise Sweep}
To demonstrate the effects of the image feature and force measurement error on the painting performance, we performed the painting task with different noise added to the measurement obtained from the sensor plugins in Gazebo simulator. Image feature measurement noise $\sigma_{vs} \in \{0.02, 0.08, 0.12\}$, which is roughly equivalent to $\{1.0, 4.0, 6.0\}\mathrm{cm}$ vehicle motion in 3D space, and force measurement noise $\sigma_f \in  \{ 1.0, 2.0, 3.0\}\mathrm{N}$. For each combination of $(\sigma_{vs}, \sigma_f)$ (in total 3), we repeated 10 times of the flights and analyzed the statistics. 

Figure~\ref{fig:statistics} shows the box plots for the aerial manipulator and the motion and force tracking accuracy. The plots show consistent tracking results in all the different noise profiles that were tested. The numeric value of the tracking error is similar to the results presented in Section~\ref{sec:demo} with the force holding achieving the accuracy within the sensor measurement noise. By observing the error statistics, we see that as the measurement noise increases, the tracking error of both force and motion increases accordingly but keeps being within the sensor noise level. 
In addition, we notice that the system struggles more with higher force tracking error at the initial and final points as well as the segments containing sharp curvature e.g. the corner where the painting direction switched. In contrast, the performance on most of the long straight line segments remains similar. The possible reason is that under sudden acceleration change, the vehicle cannot maintain perfect zero tilt. The slight tilting causes the pushing force not orthogonal to the task surface, leading to the force tracking error.


\begin{figure}
\begin{center}
\setlength{\abovecaptionskip}{0pt}
\includegraphics[width=0.9\linewidth]{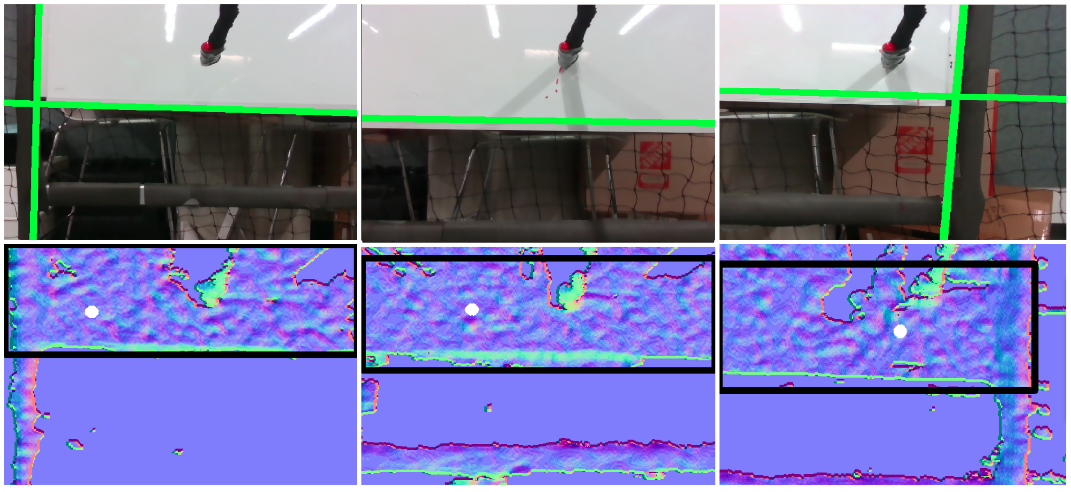}
\end{center}
\vspace{-5pt}
   \caption{Line detection performance. The top row shows three snapshots with the detected lines when the vehicle was moving from right to the left of a while board. The bottom row shows the corresponding surface normal and detected bounding box.}
   \vspace{-15pt}
\label{fig:realline}
\end{figure}

\begin{figure}
\centering
\includegraphics[width=1.0\linewidth]{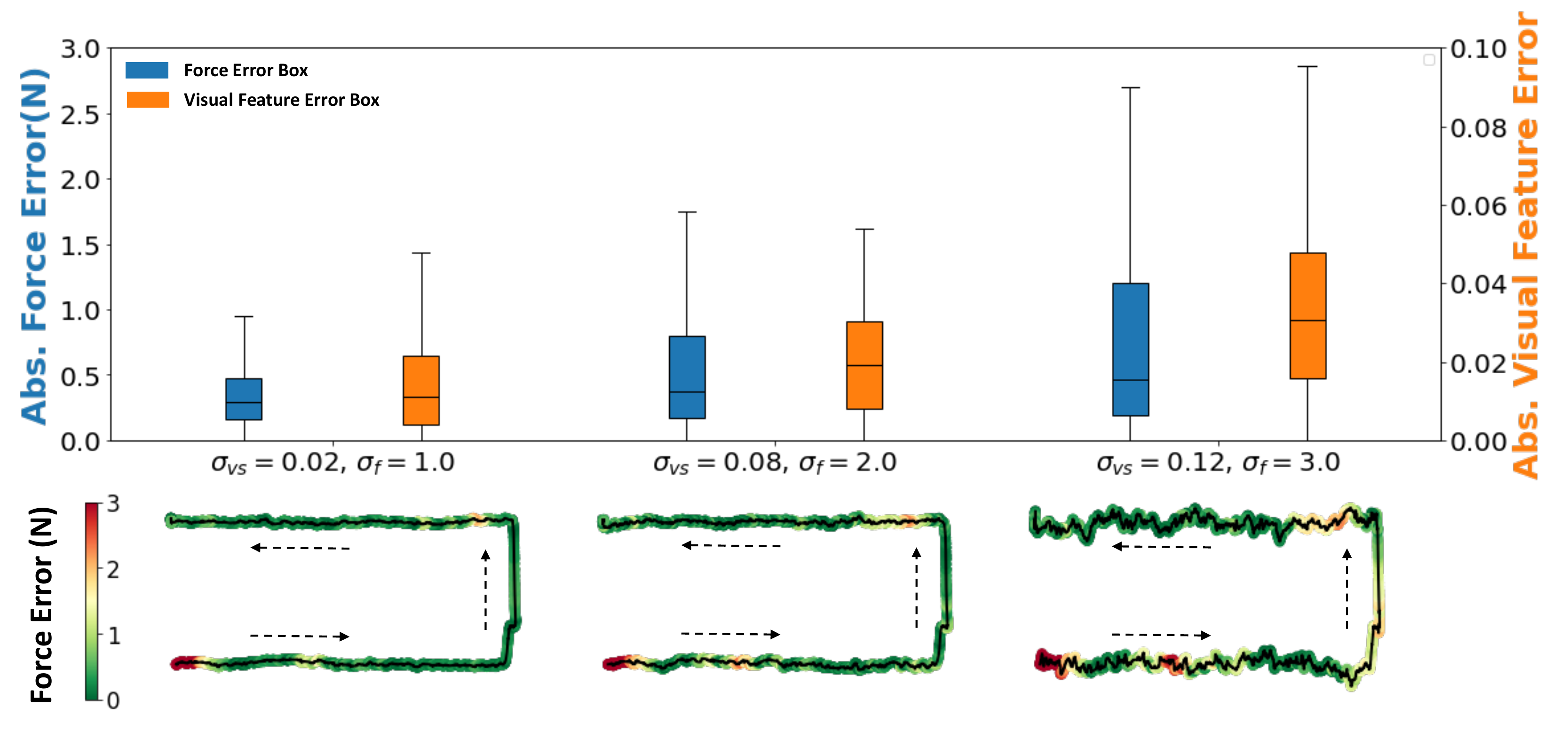}
\caption{Effects of measurement error on the painting performance. (Top) Aerial manipulator motion and force tracking box plots. (Bottom) Visual error plot for painting path under different noise profiles. (repeated 10 times). }\label{fig:statistics}
\vspace{-20pt}
\end{figure}
\section{Conclusion and Future Work} \label{sec:conclusion}
This paper develops an image-based visual servo control strategy for bridge painting using a fully-actuated UAV. The system consists of two major components: a hybrid motion and impedance force control system and a visual line detection and tracking system. Our approach does not rely on either robot pose/velocity information from an external localization system or any pre-defined visual markers. The fully-actuated UAV platform also simplifies attitude control by leveraging the zero-tilt strategy. 
Experiments show that the system can effectively execute motion tracking and force holding using only the visual guidance for the bridge painting application. Future work includes performing the system integration for more real flight tests.







\bibliographystyle{IEEEtran}
\bibliography{paper-citations.bib}

\end{document}